\documentclass[10pt,twocolumn,letterpaper]{article}

\usepackage{iccv}
\usepackage{times}
\usepackage{epsfig}
\usepackage{graphicx}
\usepackage{amsmath}
\usepackage{amssymb}
\usepackage{subfigure}
\usepackage{scrextend}

\usepackage[pagebackref=true,breaklinks=true,letterpaper=true,colorlinks,bookmarks=false]{hyperref}

\iccvfinalcopy 

\def\iccvPaperID{1768} 

\ificcvfinal\fi
\begin{document}

\title{Compression Artifacts Reduction by a Deep Convolutional Network}

\author{Chao~Dong \qquad Yubin~Deng \qquad Chen~Change~Loy \qquad Xiaoou~Tang\\
Department of Information Engineering, The Chinese University of Hong Kong\\
{\tt\small \{dc012,dy113,ccloy,xtang\}@ie.cuhk.edu.hk}
}



\maketitle
\thispagestyle{empty}

\begin{abstract}
Lossy compression introduces complex compression artifacts, particularly the blocking artifacts, ringing effects and blurring. Existing algorithms either focus on removing blocking artifacts and produce blurred output, or restores sharpened images that are accompanied with ringing effects.
Inspired by the deep convolutional networks (DCN) on super-resolution~\cite{Dong2014}, we formulate a compact and efficient network for seamless attenuation of different compression artifacts.
We also demonstrate that a deeper model can be effectively trained with the features learned in a shallow network. Following a similar ``easy to hard'' idea, we systematically investigate several practical transfer settings and show the effectiveness of transfer learning in low-level vision problems.
Our method shows superior performance than the state-of-the-arts both on the benchmark datasets and the real-world use case  (\ie~\textit{Twitter}).
In addition, we show that our method can be applied as pre-processing to facilitate other low-level vision routines when they take compressed images as input.
\end{abstract}
\section{Introduction}
\label{sec:introduction}
\begin{figure}[t]\footnotesize
\centering
\subfigure[Left: the JPEG-compressed image, where we could see blocking artifacts, ringing effects and blurring on the eyes, abrupt intensity changes on the face. Right: the restored image by the proposed deep model (AR-CNN), where we remove these compression artifacts and produce sharp details.]{
  \label{fig:introductiona}
  \includegraphics[width=\linewidth]{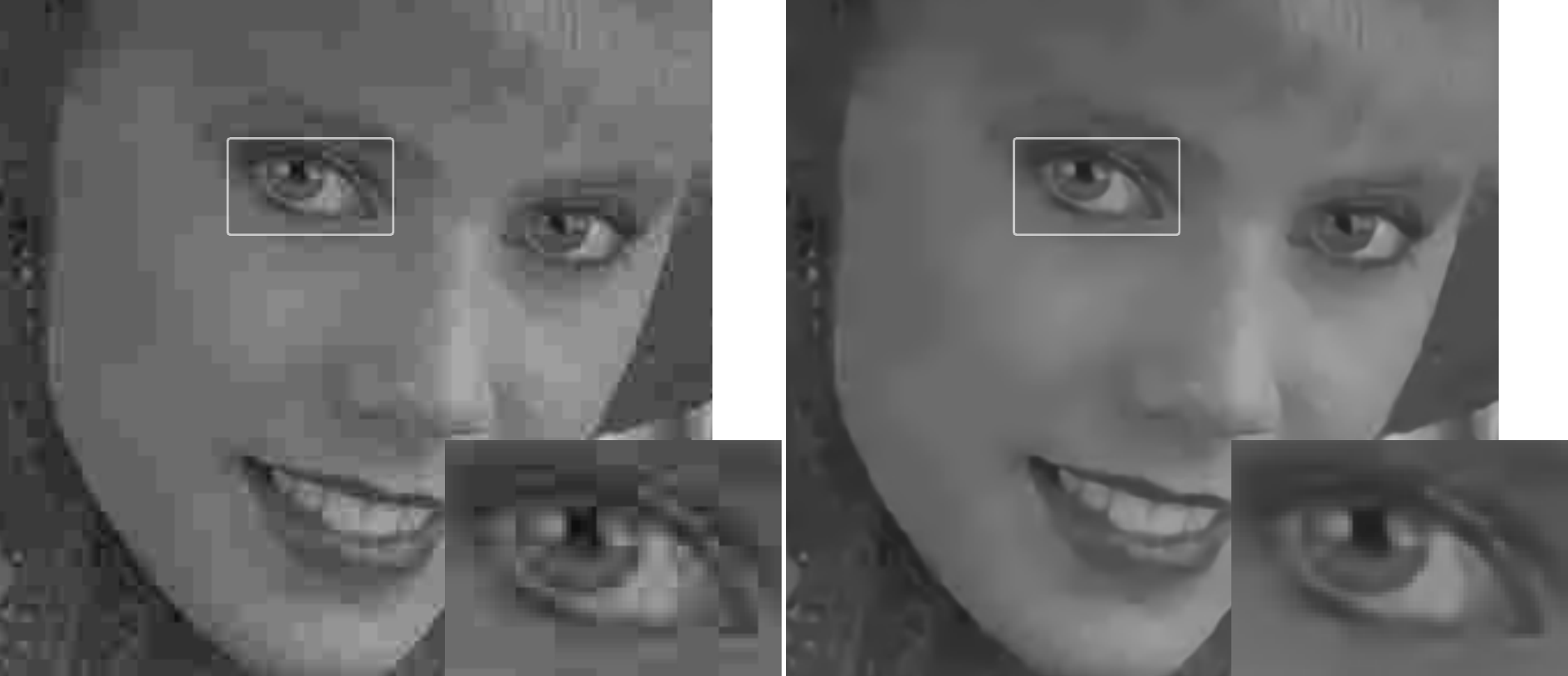}
}
\vskip -0.1cm

\subfigure[Left: the Twitter-compressed image, which is first re-scaled to a small image and then compressed on the server-side. Right: the restored image by the proposed deep model (AR-CNN)]{
  \label{fig:introductionb}
  \includegraphics[width=\linewidth]{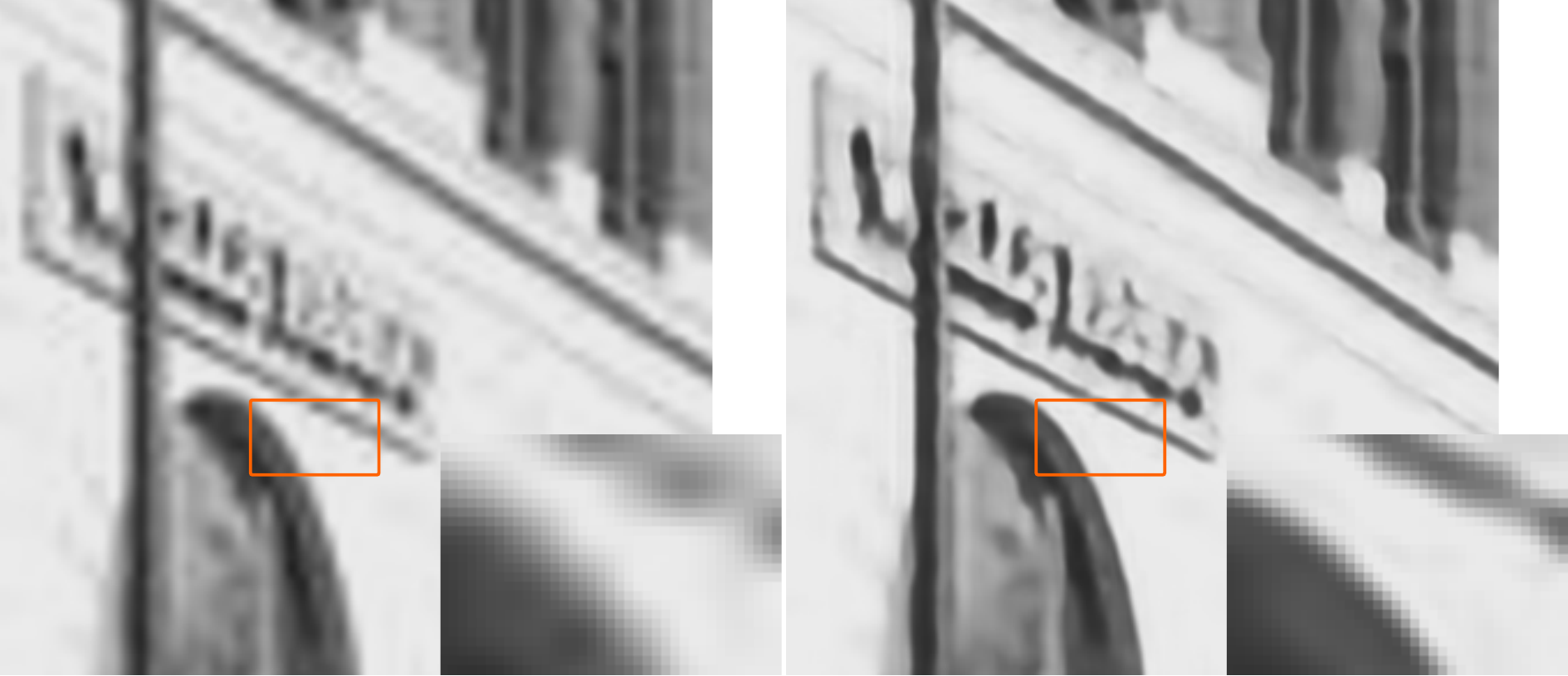}
}
\vskip -0.1cm
  \label{fig:introduction}
  \caption{Example compressed images and our restoration results on the JPEG compression scheme and the real use case -- \textit{Twitter}.}
\vskip -0.5cm
\end{figure}

Lossy compression (\eg~JPEG, WebP and HEVC-MSP) is one class of data encoding methods that uses inexact approximations for representing the encoded content. In this age of information explosion, lossy compression is indispensable and inevitable for companies (\eg~\textit{Twitter} and~\textit{Facebook}) to save bandwidth and storage space. However, compression in its nature will introduce undesired complex artifacts, which will severely reduce the user experience (\eg~Figure~\ref{fig:introduction}). All these artifacts not only decrease perceptual visual quality, but also adversely affect various low-level image processing routines that take compressed images as input, \eg~contrast enhancement~\cite{Li2014}, super-resolution~\cite{Yang2014,Dong2014}, and edge detection~\cite{Dollar2013}. However, under such a huge demand, effective compression artifacts reduction remains an open problem.

We take JPEG compression as an example to explain compression artifacts.
JPEG compression scheme divides an image into 8$\times$8 pixel blocks and applies block discrete cosine transformation (DCT) on each block individually. Quantization is then applied on the DCT coefficients to save storage space. This step will cause a complex combination of different artifacts, as depicted in Figure~\ref{fig:introductiona}.
\textit{Blocking artifacts} arise when each block is encoded without considering the correlation with the adjacent blocks, resulting in discontinuities at the 8$\times$8 borders.
\textit{Ringing effects} along the edges occur due to the coarse quantization of the high-frequency components (also known as Gibbs phenomenon~\cite{Gonzalez2002}).
\textit{Blurring} happens due to the loss of high-frequency components.
To cope with the various compression artifacts, different approaches have been proposed, some of which can only deal with certain types of artifacts.
For instance, deblocking oriented approaches~\cite{List2003,ReeveIII1984,Wang2013} perform filtering along the block boundaries to reduce only blocking artifacts. Liew~\etal~\cite{Liew2004} and Foi~\etal~\cite{Foi2007} use thresholding by wavelet transform and Shape-Adaptive DCT transform, respectively. These approaches are good at removing blocking and ringing artifacts, but tend to produce blurred output. Jung~\etal~\cite{Jung2012} propose restoration method based on sparse representation. They produce sharpened images but accompanied with noisy edges and unnatural smooth regions. 

To date, deep learning has shown impressive results on both high-level and low-level vision problems . In particular, the SRCNN proposed by Dong~\etal~\cite{Dong2014} shows the great potential of an end-to-end DCN in image super-resolution. The study also points out that conventional sparse-coding-based image restoration model can be equally seen as a deep model. However, we find that the three-layer network is not well suited in restoring the compressed images, especially in dealing with blocking artifacts and handling smooth regions. As various artifacts are coupled together, features extracted by the first layer is noisy, causing undesirable noisy patterns in reconstruction.

To eliminate the undesired artifacts, we improve the SRCNN by embedding one or more ``feature enhancement'' layers after the first layer to clean the noisy features. Experiments show that the improved model, namely ``Artifacts Reduction Convolutional Neural Networks (AR-CNN)'', is exceptionally effective in suppressing blocking artifacts while retaining edge patterns and sharp details (see Figure~\ref{fig:introduction}). However, we are met with training difficulties in training a deeper DCN. ``Deeper is better'' is widely observed in high-level vision problems, but not in low-level vision tasks. Specifically, ``deeper is not better'' has been pointed out in super-resolution~\cite{Dong2014a}, where training a five-layer network becomes a bottleneck. The difficulty of training is partially due to the sub-optimal initialization settings.

The aforementioned difficulty motivates us to investigate a better way to train a deeper model for low-level vision problems. We find that this can be effectively solved by transferring the features learned in a shallow network to a deeper one and fine-tuning simultaneously\footnote{Generally, the transfer learning method will train a base network first, and copy the learned parameters or features of several layers to the corresponding layers of a target network. These transferred layers can be left frozen or fine-tuned to the target dataset. The remaining layers are randomly initialized and trained to the target task.}. This strategy has also been proven successful in learning a deeper CNN for image classification~\cite{Simonyan2014}.
Following a similar general intuitive idea, \textit{easy to hard}, we discover other interesting transfer settings in this low-level vision task:
(1) We transfer the features learned in a high-quality compression model (easier) to a low-quality one (harder), and find that it converges faster than random initialization.
(2) In the real use case, companies tend to apply different compression strategies (including re-scaling) according to their purposes (\eg~Figure~\ref{fig:introductionb}). We transfer the features learned in a standard compression model (easier) to a real use case (harder), and find that it performs better than learning from scratch.

The contributions of this study are three-fold:
(1) We formulate a new deep convolutional network for efficient reduction of various compression artifacts.
Extensive experiments, including that on real use cases,
demonstrate the effectiveness of our method over state-of-the-art methods~\cite{Foi2007,Jancsary2012} both perceptually and quantitatively.
(2) We verify that reusing the features in shallow networks is helpful in learning a deeper model for compression artifact reduction. Under the same intuitive idea -- \textit{easy to hard}, we reveal a number of interesting and practical transfer settings. Our study is the first attempt to show the effectiveness of feature transfer in a low-level vision problem.
(3) We show the effectiveness of AR-CNN in facilitating other low-level vision routines (\ie~super-resolution and contrast enhancement), when they take JPEG images as input.

\begin{figure*}[t]\small
\begin{center}

 \includegraphics[width=0.96\linewidth]{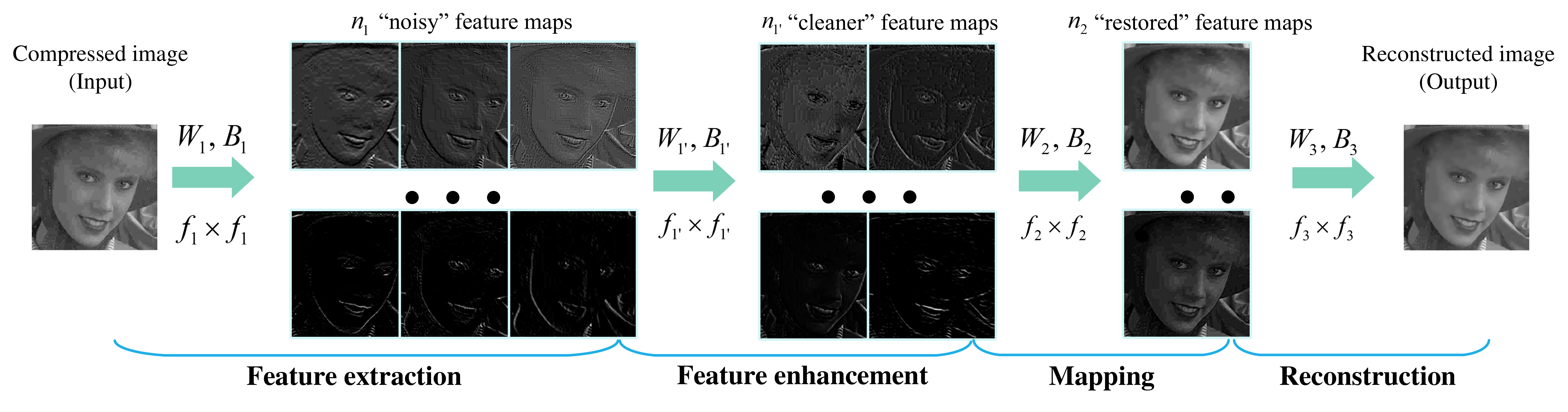}
\caption{The framework of the Artifacts Reduction Convolutional Neural Network (AR-CNN). The network consists of four convolutional layers, each of which is responsible for a specific operation. Then it optimizes the four operations (\ie,~feature extraction, feature enhancement, mapping and reconstruction) jointly in an end-to-end framework. Example feature maps shown in each step could well illustrate the functionality of each operation. They are normalized for better visualization.}
\label{fig:framework}
\vspace{-0.6cm}
\end{center}
\end{figure*}

\section{Related work}

Existing algorithms can be classified into deblocking oriented and restoration oriented methods.
The deblocking oriented methods focus on removing blocking and ringing artifacts.
In the spatial domain, different kinds of filters~\cite{List2003,ReeveIII1984,Wang2013} have been proposed to adaptively deal with blocking artifacts in specific regions (\eg,~edge, texture, and smooth regions). In the frequency domain, Liew~\etal~\cite{Liew2004} utilize wavelet transform and derive thresholds at different wavelet scales for denoising. The most successful deblocking oriented method is perhaps the Pointwise Shape-Adaptive DCT (SA-DCT)~\cite{Foi2007}, which is widely acknowledged as the state-of-the-art approach~\cite{Jancsary2012,Li2014}.
However, as most deblocking oriented methods, SA-DCT could not reproduce sharp edges, and tend to overly smooth texture regions.
The restoration oriented methods regard the compression operation as distortion and propose restoration algorithms. They include projection on convex sets based method (POCS)~\cite{Yang1995}, solving an MAP problem (FoE)~\cite{Sun2007}, sparse-coding-based method~\cite{Jung2012} and the Regression Tree Fields based method (RTF)~\cite{Jancsary2012}, which is the new state-of-the art method. The RTF takes the results of SA-DCT~\cite{Foi2007} as bases and produces globally consistent image reconstructions with a regression tree field model.
It could also be optimized for any differentiable loss functions (\eg~SSIM), but often at the cost of other evaluation metrics. 

Super-Resolution Convolutional Neural Network (SRCNN)~\cite{Dong2014} is closely related to our work. In the study, independent steps in the sparse-coding-based method are formulated as different convolutional layers and optimized in a unified network. It shows the potential of deep model in low-level vision problems like super-resolution. However, the model of compression is different from super-resolution in that it consists of different kinds of artifacts. Designing a deep model for compression restoration requires a deep understanding into the different artifacts. We show that directly applying the SRCNN architecture for compression restoration will result in undesired noisy patterns in the reconstructed image. 

Transfer learning in deep neural networks becomes popular since the success of deep learning in image classification~\cite{Krizhevsky2012}. The features learned from the ImageNet show good generalization ability~\cite{Zeiler2014} and become a powerful tool for several high-level vision problems, such as Pascal VOC image classification~\cite{Oquab2014} and object detection~\cite{Girshick2014,Sermanet2013}.
Yosinski~\etal~\cite{Yosinski2014} have also tried to quantify the degree to which a particular layer is general or specific. Overall, transfer learning has been systematically investigated in high-level vision problems, but not in low-level vision tasks. In this study, we explore several transfer settings on compression artifacts reduction and show the effectiveness of transfer learning in low-level vision problems.

\section{Methodology}
\label{sec:Methodology}

Our proposed approach is based on the current successful low-level vision model -- SRCNN~\cite{Dong2014}. To have a better understanding of our work, we first give a brief overview of SRCNN. Then we explain the insights that lead to a deeper network and present our new model. Subsequently, we explore three types of transfer learning strategies that help in training a deeper and better network.

\subsection{Review of SRCNN}
\label{sec:SRCNN}
The SRCNN aims at learning an end-to-end mapping, which takes the low-resolution image $\mathbf{Y}$ (after interpolation) as input and directly outputs the high-resolution one $F(\mathbf{Y})$. The network contains three convolutional layers, each of which is responsible for a specific task. Specifically, the first layer performs \textbf{patch extraction and representation}, which extracts overlapping patches from the input image and represents each patch as a high-dimensional vector. Then the \textbf{non-linear mapping} layer maps each high-dimensional vector of the first layer to another high-dimensional vector, which is conceptually the representation of a high-resolution patch. At last, the \textbf{reconstruction} layer aggregates the patch-wise representations to generate the final output.
The network can be expressed as:
\begin{align}
\label{eqn:SRCNN}
F_{i}(\mathbf{Y})&=\max\left(0, W_{i}*\mathbf{Y}+B_{i}\right), i\in\{1,2\}; \\ F(\mathbf{Y})&=W_3*F_{2}(\mathbf{Y})+B_3.
\end{align}
where $W_{i}$ and $B_{i}$ represent the filters and biases of the $i$th layer respectively, $F_{i}$ is the output feature maps and '$*$' denotes the convolution operation. The $W_{i}$ contains $n_i$ filters of support $n_{i-1}\times f_i \times f_i$, where $f_i$ is the spatial support of a filter, $n_i$ is the number of filters, and $n_0$ is the number of channels in the input image. Note that there is no pooling or full-connected layers in SRCNN, so the final output $F(\mathbf{Y})$ is of the same size as the input image.
Rectified Linear Unit (ReLU, $\max(0,x)$)~\cite{Nair2010} is applied on the filter responses.

These three steps are analogous to the basic operations in the sparse-coding-based super-resolution methods~\cite{Yang2010a}, and this close relationship lays theoretical foundation for its successful application in super-resolution. Details can be found in the paper~\cite{Dong2014}.

\subsection{Convolutional Neural Network for Compression Artifacts Reduction}
\label{sec:ARCNN}
\textbf{Insights.} In sparse-coding-based methods and SRCNN, the first step -- feature extraction -- determines what should be emphasized and restored in the following stages. However, as various compression artifacts are coupled together, the extracted features are usually noisy and ambiguous for accurate mapping. In the experiments of reducing JPEG compression artifacts (see Section~\ref{sec:exp_SRCNN}), we find that some quantization noises coupled with high frequency details are further enhanced, bringing unexpected noisy patterns around sharp edges. Moreover, blocking artifacts in flat areas are misrecognized as normal edges, causing abrupt intensity changes in smooth regions. Inspired by the feature enhancement step in super-resolution~\cite{Xiong2009}, we introduce a feature enhancement layer after the feature extraction layer in SRCNN to form a new and deeper network -- AR-CNN. This layer maps the ``noisy'' features to a relatively ``cleaner'' feature space, which is equivalent to denoising the feature maps.

\textbf{Formulation.} The overview of the new network AR-CNN is shown in Figure~\ref{fig:framework}. The three layers of SRCNN remain unchanged in the new model. We also use the same annotations as in Section~\ref{sec:SRCNN}. To conduct feature enhancement,
we extract new features from the $n_1$ feature maps of the first layer, and combine them to form another set of feature maps.
This operation $F_{1'}$ can also be formulated as a convolutional layer:
\begin{equation}
\label{eqn:first_layer}
F_{1'}(\mathbf{Y})=\max\left(0, W_{1'}*F_1(\mathbf{Y})+B_{1'}\right),
\end{equation}
where $W_{1'}$ corresponds to $n_{1'}$ filters with size $n_1\times f_{1'}\times f_{1'}$. $B_{1'}$ is an $n_{1'}$-dimensional bias vector, and the output $F_{1'}(\mathbf{Y})$ consists of $n_{1'}$ feature maps. Overall, the AR-CNN consists of four layers, namely the feature extraction, feature enhancement, mapping and reconstruction layer.

It is worth noticing that AR-CNN is not equal to a deeper SRCNN that contains more than one non-linear mapping layers\footnote{Adding non-linear mapping layers has been suggested as an extension of SRCNN in~\cite{Dong2014}.}. Rather than imposing more non-linearity in the mapping stage, AR-CNN improves the mapping accuracy by enhancing the extracted low-level features. Experimental results of AR-CNN, SRCNN and deeper SRCNN will be shown in Section~\ref{sec:exp_SRCNN}

\subsection{Model Learning}
\label{subsec:learning}
Given a set of ground truth images $\left\{ \mathbf{X}_i\right\}$ and their corresponding compressed images $\left\{ \mathbf{Y}_i\right\}$, we use Mean Squared Error (MSE) as the loss function:
\begin{equation}
\label{eqn:loss}
L(\Theta)=\frac{1}{n}\sum_{i=1}^n||F(\mathbf{Y}_i ; \Theta) - \mathbf{X}_i||^2,
\end{equation}
where $\Theta=\{W_1,W_{1'},W_2,W_3,B_1,B_{1'},B_2,B_3\}$, $n$ is the number of training samples. The loss is minimized using stochastic gradient descent with the standard backpropagation.
We adopt a batch-mode learning method with a batch size of 128.

\subsection{Easy-Hard Transfer}

\begin{figure}[t]\small
\begin{center}
\vskip -0.6cm
 \includegraphics[width=1\linewidth]{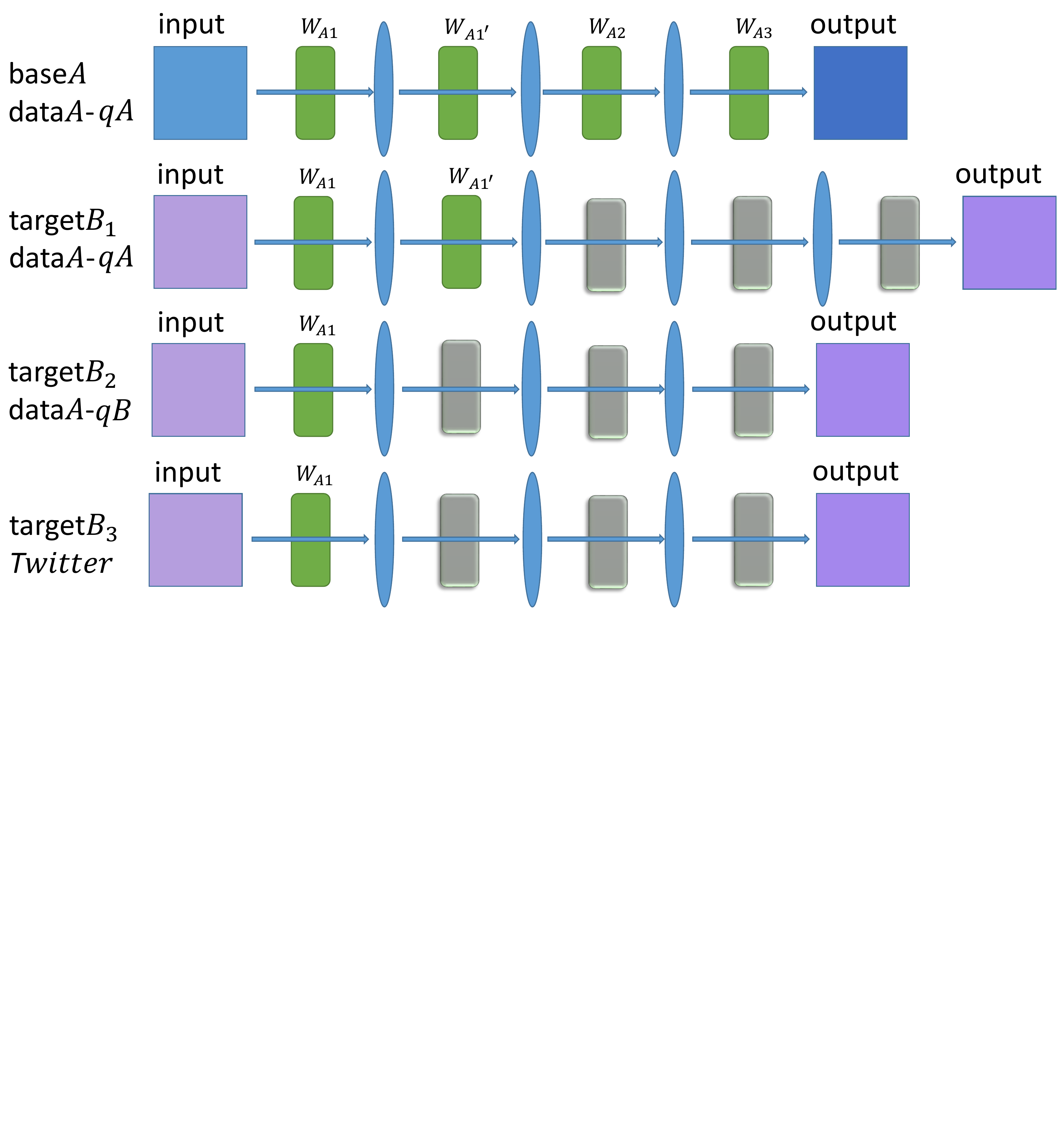}
\vskip -3.9cm
\caption{Easy-hard transfer settings. First row: The baseline 4-layer network trained with \textit{dataA}-$qA$. Second row: The 5-layer AR-CNN targeted at \textit{dataA}-$qA$. Third row: The AR-CNN targeted at \textit{dataA}-$qB$. Fourth row: The AR-CNN targeted at \textit{Twitter} data. Green boxes indicate the transferred features from the base network, and gray boxes represent random initialization. The ellipsoidal bars between weight vectors represent the activation functions.}
\label{fig:Easy-hard1}
\vspace{-0.75cm}
\end{center}
\end{figure}

Transfer learning in deep models provides an effective way of initialization. In fact, conventional initialization strategies (\ie~randomly drawn from Gaussian distributions with fixed standard deviations~\cite{Krizhevsky2012}) are found not suitable for training a very deep model, as reported in~\cite{He2015}.
To address this issue, He~\etal~\cite{He2015} derive a robust initialization method for rectifier nonlinearities,  Simonyan~\etal~\cite{Simonyan2014} propose to use the pre-trained features on a shallow network for initialization.

In low-level vision problems (\eg~super resolution), it is observed that training a network beyond 4 layers would encounter the problem of convergence, even that a large number of training images (\eg~ImageNet) are provided~\cite{Dong2014}.
We are also met with this difficulty during the training process of AR-CNN.
To this end, we systematically investigate several transfer settings in training a low-level vision network following an intuitive idea of ``easy-hard transfer''.
Specifically, we attempt to reuse the features learned in a relatively easier task to initialize a deeper or harder network.
Interestingly, the concept ``easy-hard transfer'' has already been pointed out in neuro-computation study~\cite{Gluck1993}, where the prior training on an easy discrimination can help learn a second harder one.

Formally, we define the base (or source) task as \textit{A} and the target tasks as \textit{$B_i$}, $i\in \{1,2,3\}$. As shown in Figure~\ref{fig:Easy-hard1}, the base network \textit{baseA} is a four-layer AR-CNN trained on a large dataset \textit{dataA}, of which images are compressed using a standard compression scheme with the compression quality \textit{qA}. All layers in \textit{baseA} are randomly initialized from a Gaussian distribution. We will transfer one or two layers of \textit{baseA} to different target tasks (see Figure~\ref{fig:Easy-hard1}). Such transfers can be described as follows.

\textbf{Transfer shallow to deeper model.}
As indicated by~\cite{Dong2014a}, a five-layer network is sensitive to the initialization parameters and learning rate. Thus we transfer the first two layers of \textit{baseA} to a five-layer network \textit{target$B_1$}. Then we randomly initialize its remaining layers\footnote{Random initialization on remaining layers are also applied similarly for tasks \textit{$B_2$}, and \textit{$B_3$}.} and train all layers toward the same dataset \textit{dataA}. This is conceptually similar to that applied in image classification~\cite{Simonyan2014}, but this approach has never been validated in low-level vision problems.

\textbf{Transfer high to low quality.}
Images of low compression quality contain more complex artifacts. Here we use the features learned from high compression quality images as a starting point to help learn more complicated features in the DCN. Specifically, the first layer of \textit{target$B_2$} are copied from \textit{baseA} and trained on images that are compressed with a lower compression quality \textit{qB}.

\textbf{Transfer standard to real use case.}
We then explore whether the features learned under a standard compression scheme can be generalized to other real use cases, which often contain more complex artifacts due to different levels of re-scaling and compression. We transfer the first layer of \textit{baseA} to the network \textit{target$B_3$}, and train all layers on the new dataset.

\begin{figure}[t]\tiny
\centering
\subfigure[High compression quality (quality 20 in Matlab encoder)]{
  \label{pattern_q20}
  \includegraphics[width=\linewidth]{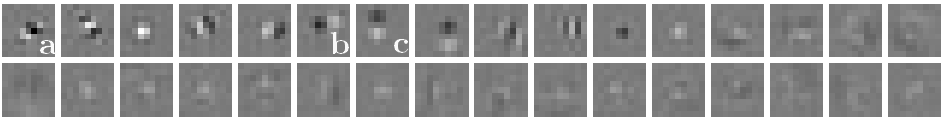}
}
\vskip -0.2cm
\subfigure[Low compression quality (quality 10 in Matlab encoder)]{
  \label{pattern_q10}
  \includegraphics[width=\linewidth]{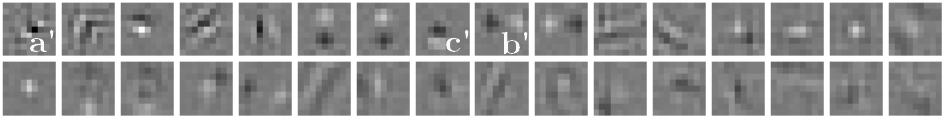}
}
\vskip -0.16cm
\label{fig:features}
  \caption{First layer filters of AR-CNN learned under different JPEG compression qualities.}
\vskip -0.4cm
\end{figure}

\textbf{Discussion.}
Why the features learned from relatively easy tasks are helpful? First, the features from a well-trained network can provide a good starting point. Then the rest of a deeper model can be regarded as shallow one, which is easier to converge. Second, features learned in different tasks always have a lot in common. For instance, Figure~\ref{fig:features} shows the features learned under different JPEG compression qualities. Obviously, filters $a, b, c$ of high quality are very similar to filters $a', b', c'$ of low quality. This kind of features can be reused or improved during fine-tuning, making the convergence faster and more stable. Furthermore, a deep network for a hard problem can be seen as an insufficiently biased learner with overly large hypothesis space to search, and therefore is prone to overfitting. These few transfer settings we investigate introduce good bias to enable the learner to acquire a concept with greater generality. Experimental results in Section~\ref{sec:transfer} validate the above analysis.

\section{Experiments}
\label{subsec:settings}
We use the BSDS500 database~\cite{Arbelaez2011} as our base training set. Specifically, its disjoint training set (200 images) and test set (200 images) are all used for training, and its validation set (100 images) is used for validation.
As in other compression artifacts reduction methods (\eg~RTF~\cite{Jancsary2012}), we apply the standard JPEG compression scheme, and use the JPEG quality settings $q=20$~(mid quality) and $q=10$~(low quality) in MATLAB JPEG encoder. We only focus on the restoration of the luminance channel (in YCrCb space) in this paper.

The training image pairs $\left\{Y, X \right\}$ are prepared as follows -- Images in the training set are decomposed into $32\times 32$ sub-images\footnote{We use sub-images because we regard each sample as an image rather than a big patch.} $X = \{\mathbf{X}_i\}_{i=1}^n$. Then the compressed samples $Y=\{\mathbf{Y}_i\}_{i=1}^n$ are generated from the training samples with MATLAB JPEG encoder~\cite{Jancsary2012}. The sub-images are extracted from the ground truth images with a stride of 10. Thus the 400 training images could provide 537,600 training samples. To avoid the border effects caused by convolution, AR-CNN produces a $20\times 20$ output given a $32 \times 32$ input $\mathbf{Y}_i$. Hence, the loss (Eqn.~(\ref{eqn:loss})) was computed by comparing against the center $20\times 20$ pixels of the ground truth sub-image $\mathbf{X}_i$.
In the training phase, we follow~\cite{Jain2009,Dong2014} and use a smaller learning rate ($10^{-5}$) in the last layer and a comparably larger one ($10^{-4}$) in the remaining layers.

\subsection{Comparison with the State-of-the-Arts}

We use the LIVE1 dataset~\cite{Sheikh2005} (29 images) as test set to evaluate both the quantitative and qualitative performance. The LIVE1 dataset contains images with diverse properties. It is widely used in image quality assessment~\cite{Wang2004} as well as in super-resolution~\cite{Yang2014}.
To have a comprehensive qualitative evaluation, we apply the PSNR, structural similarity (SSIM)~\cite{Wang2004}\footnote{We use the unweighted structural similarity defined over fixed $8\times 8$ windows as in~\cite{Wang2008}.}, and PSNR-B~\cite{Yim2011} for quality assessment. We want to emphasize the use of PSNR-B. It is designed specifically to assess blocky and deblocked images, thus is more sensitive to blocking artifacts than the perceptual-aware SSIM index.
The network settings are $f_1=9$, $f_{1'}=7$, $f_2=1$, $f_3=5$, $n_1=64$, $n_{1'}=32$, $n_2=16$ and $n_3=1$, denoted as AR-CNN (9-7-1-5) or simply AR-CNN. A specific network is trained for each JPEG quality. Parameters are randomly initialized from a Gaussian distribution with a standard deviation of 0.001.

\subsubsection{Comparison with SA-DCT}

\label{sec:sadct}
\begin{table}\scriptsize
\caption{The average results of PSNR (dB), SSIM, PSNR-B (dB) on the LIVE1 dataset.}\label{tab:sadct}
\vspace{-0.15cm}
\begin{center}
\begin{tabular}{|c|c|c|c|c|}
\hline
 Eval. Mat & Quality & JPEG & SA-DCT & AR-CNN \\

\hline\hline
PSNR & 10 & 27.77 & 28.65 & \textbf{28.98}   \\
     & 20 & 30.07 & 30.81 & \textbf{31.29}   \\
\hline\hline
SSIM & 10 & 0.7905 & 0.8093 & \textbf{0.8217}   \\
     & 20 & 0.8683 & 0.8781 & \textbf{0.8871}   \\
\hline\hline
PSNR-B & 10 & 25.33 & 28.01 & \textbf{28.70}   \\
     & 20 & 27.57 & 29.82 & \textbf{30.76}  \\
\hline
\end{tabular}
\vspace{-0.65cm}
\end{center}
\end{table}

We first compare AR-CNN with SA-DCT~\cite{Foi2007}, which is widely regarded as the state-of-the-art deblocking oriented method~\cite{Jancsary2012,Li2014}. The quantization results of PSNR, SSIM and PSNR-B are shown in Table~\ref{tab:sadct}. On the whole, our AR-CNN outperforms the SA-DCT on all JPEG qualities and evaluation metrics by a large margin. Note that the gains on PSNR-B is much larger than that on PSNR. This indicates that AR-CNN could produce images with less blocking artifacts.
To compare the visual quality, we present some restored images\footnote{\label{note1}More qualitative results are provided in the supplementary file.} with $q=10$ in Figure~\hyperlink{page.8}{\textcolor{red}{10}}.
From Figure~\hyperlink{page.8}{\textcolor{red}{10}}, we could see that the result of AR-CNN could produce much sharper edges with much less blocking and ringing artifacts compared with SA-DCT.
The visual quality has been largely improved on all aspects compared with the state-of-the-art method.
Furthermore, AR-CNN is superior to SA-DCT on the implementation speed. For SA-DCT, it needs 3.4 seconds to process a $256\times 256$ image. While AR-CNN only takes 0.5 second. They are all implemented using C++ on a PC with Intel I3 CPU (3.1GHz) with 16GB RAM.

\subsubsection{Comparison with SRCNN}
\label{sec:exp_SRCNN}

\begin{table}\scriptsize
\caption{The average results of PSNR (dB), SSIM, PSNR-B (dB) on the LIVE1 dataset with $q=10$ .}\label{tab:srcnn}
\vspace{-0.15cm}
\begin{center}
\begin{tabular}{|c|c|c|c|c|}
\hline
 Eval. & JPEG & SRCNN  & Deeper & AR-CNN \\
 Mat &       &         & SRCNN &    \\

\hline\hline
PSNR  & 27.77 & 28.91 & 28.92 & \textbf{28.98}   \\
\hline
SSIM & 0.7905 & 0.8175 & 0.8189 & \textbf{0.8217}   \\
\hline
PSNR-B  & 25.33 & 28.52 & 28.46 & \textbf{28.70}   \\
\hline
\end{tabular}
\vspace{-0.25cm}
\end{center}
\end{table}

\begin{figure}[t]\small
\begin{center}
 \includegraphics[width=\linewidth]{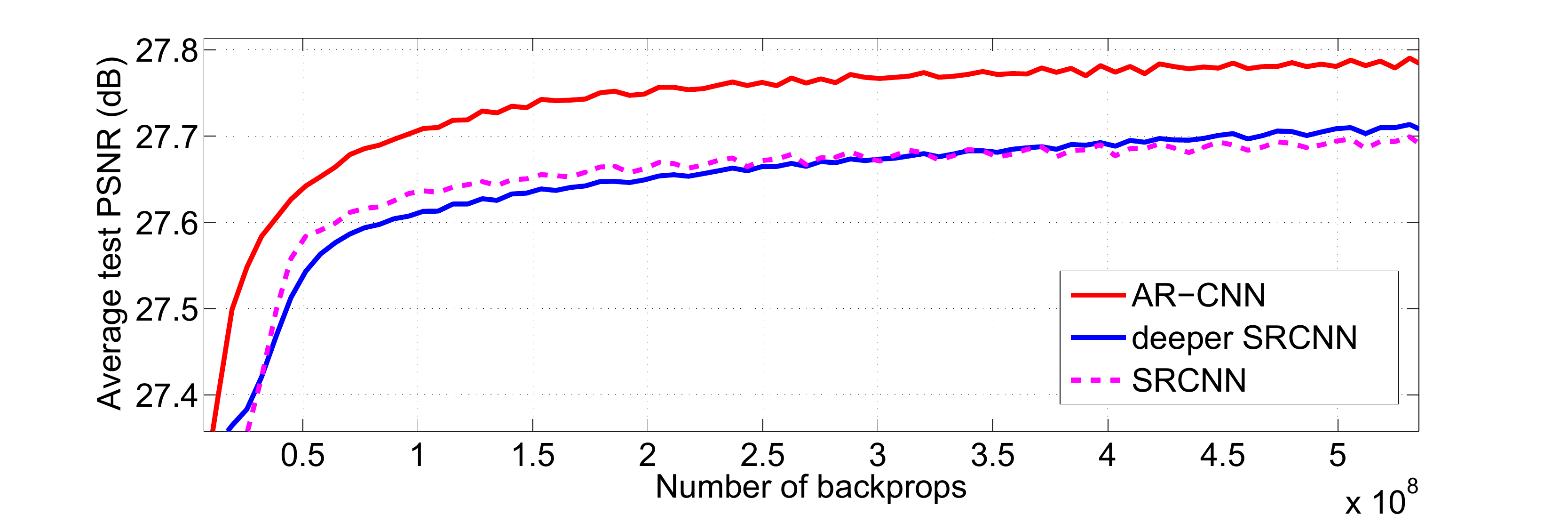}
\caption{Comparisons with SRCNN and Deeper SRCNN.}
\label{fig:srcnn}
\vspace{-0.75cm}
\end{center}
\end{figure}

As discussed in Section~\ref{sec:ARCNN}, SRCNN is not suitable for compression artifacts reduction. For comparison, we train two SRCNN networks with different settings.
(i) The original SRCNN (9-1-5) with $f_1=9$, $f_3=5$, $n_1=64$ and $n_2=32$. (ii) Deeper SRCNN (9-1-1-5) with an additional non-linear mapping layer ($f_{2'}=1$, $n_{2'}=16$).
They all use the BSDS500 dataset for training and validation as in Section~\ref{subsec:settings}. The compression quality is $q=10$. The AR-CNN is the same as in Section~\ref{sec:sadct}.

Quantitative results tested on LIVE1 dataset are shown in Table~\ref{tab:srcnn}. We could see that the two SRCNN networks are inferior on all evaluation metrics. From convergence curves shown in Figure~\ref{fig:srcnn}, it is clear that AR-CNN achieves higher PSNR from the beginning of the learning stage.
Furthermore, from their restored images\footref{note1} in Figure~\hyperlink{page.8}{\textcolor{red}{11}}, we find out that the two SRCNN networks all produce images with noisy edges and unnatural smooth regions.
These results demonstrate our statements in Section~\ref{sec:ARCNN}. In short, the success of training a deep model needs comprehensive understanding of the problem and careful design of the model structure.

\subsubsection{Comparison with RTF}

\begin{table}\scriptsize
\caption{The average results of PSNR (dB), SSIM, PSNR-B (dB) on the test set BSDS500 dataset.}\label{tab:rtf}
\vspace{-0.15cm}
\begin{center}
\begin{tabular}{|c|c|c|c|c|c|}
\hline
 Eval. & Quality & JPEG & RTF & RTF & AR-CNN \\
Mat &            &      &     & +SA-DCT&    \\
\hline\hline
PSNR & 10 & 26.62 & 27.66 & 27.71 & \textbf{27.71}   \\
     & 20 & 28.80 & 29.84 & 29.87 &\textbf{29.87}   \\
\hline\hline
SSIM & 10 & 0.7904 & 0.8177 & 0.8186 & \textbf{0.8192}   \\
     & 20 & 0.8690 & 0.8864 & \textbf{0.8871} & 0.8857   \\
\hline\hline
PSNR-B & 10 & 23.54 & 26.93 & 26.99 & \textbf{27.04}   \\
     & 20 & 25.62 & 28.80 & 28.80 & \textbf{29.02}  \\
\hline
\end{tabular}
\vspace{-0.65cm}
\end{center}
\end{table}

RTF~\cite{Jancsary2012} is the recent state-of-the-art restoration oriented method. Without their deblocking code, we can only compare with the released deblocking results. Their model is trained on the training set (200 images) of the BSDS500 dataset, but all images are down-scaled by a factor of 0.5~\cite{Jancsary2012}. To have a fair comparison, we also train new AR-CNN networks on the same half-sized 200 images. Testing is performed on the test set of the BSDS500 dataset (images scaled by a factor of 0.5), which is also consistent with~\cite{Jancsary2012}. We compare with two RTF variants. One is the plain RTF, which uses the filter bank and is optimized for PSNR. The other is the RTF+SA-DCT, which includes the SA-DCT as a base method and is optimized for MAE. The later one achieves the highest PSNR value among all RTF variants~\cite{Jancsary2012}.

As shown in Table~\ref{tab:rtf}, we obtain superior performance than the plain RTF, and even better performance than the combination of RTF and SA-DCT, especially under the more representative PSNR-B metric.
Moreover, training on such a small dataset has largely restricted the ability of AR-CNN. The performance of AR-CNN will further improve given more training images.

\subsection{Experiments on Easy-Hard Transfer}
\label{sec:transfer}
We show the experimental results of different ``easy-hard transfer'' settings, of which the details are shown in Table~\ref{tab:transfer}. Take the base network as an example, the base-q10 is a four-layer AR-CNN (9-7-1-5) trained on the BSDS500~\cite{Arbelaez2011} dataset (400 images) under the compression quality $q=10$. Parameters are initialized by randomly drawing from a Gaussian distribution with zero mean and standard deviation 0.001. Figures~\ref{fig:transfer1}~-~\ref{fig:transfer3} show the convergence curves on the validation set.

\begin{table}\scriptsize
\caption{Experimental settings of ``easy-hard transfer''.}\label{tab:transfer}
\vspace{-0.15cm}
\begin{center}
\begin{tabular}{|c|c|c|c|c|c|}
\hline
transfer & short & network   & training  & initialization \\
strategy & form  & structure & dataset   & strategy       \\
\hline\hline
base     & base-q10 & 9-7-1-5 & BSDS-q10  & Gaussian (0, 0.001)  \\
network  & base-q20 & 9-7-1-5 & BSDS-q20  & Gaussian (0, 0.001)  \\
\hline\hline
shallow  & base-q10 & 9-7-1-5 & BSDS-q10  & Gaussian (0, 0.001)  \\
to       & transfer deeper & 9-7-3-1-5 & BSDS-q10  & 1,2 layers of base-q10\\
deep    & He~\cite{He2015} & 9-7-3-1-5 & BSDS-q10  & He~\etal~\cite{He2015}\\
\hline\hline
high  & base-q10 & 9-7-1-5 &BSDS-q10  & Gaussian (0, 0.001)  \\
to       & transfer 1 layer & 9-7-1-5 & BSDS-q10  & 1 layer of base-q20\\
low   & transfer 2 layers & 9-7-1-5 &BSDS-q10  & 1,2 layer of base-q20 \\
\hline\hline
standard  & base-Twitter & 9-7-1-5 & \textit{Twitter}  & Gaussian (0, 0.001)  \\
to       & transfer q10 & 9-7-1-5 & \textit{Twitter}  & 1 layer of base-q10\\
real   & transfer q20 & 9-7-1-5 & \textit{Twitter} & 1 layer of base-q20 \\
\hline
\end{tabular}
\vspace{-0.65cm}
\end{center}
\end{table}

\subsubsection{Transfer shallow to deeper model}
\label{sec:transfer1}

\begin{figure}[t]\small
\begin{center}
 \includegraphics[width=\linewidth]{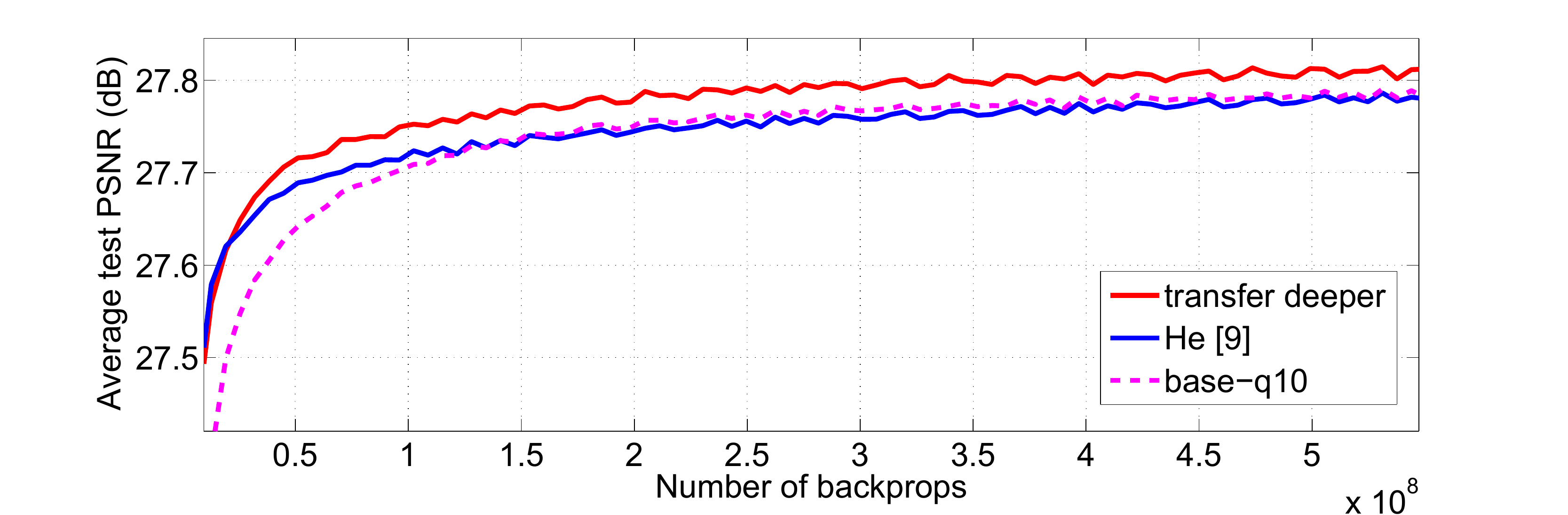}
\vskip -0.15cm
\caption{Transfer shallow to deeper model.}
\label{fig:transfer1}
\vspace{-0.75cm}
\end{center}
\end{figure}

In Table~\ref{tab:transfer}, we denote a deeper (five-layer) AR-CNN as ``9-7-3-1-5'',  which contains another feature enhancement layer ($f_{1''}=3$ and $n_{1''}=16$).
Results in Figure~\ref{fig:transfer1} show that the transferred features from a four-layer network enable us to train a five-layer network successfully. Note that directly training a five-layer network using conventional initialization ways is unreliable. Specifically, we have exhaustively tried different groups of learning rates, but still have not observed convergence. Furthermore, the ``transfer deeper'' converges faster and achieves better performance than using He~\etal's method~\cite{He2015}, which is also very effective in training a deep model. We have also conducted comparative experiments with the structure ``9-7-1-1-5'' and observed the same trend.

\subsubsection{Transfer high to low quality}
\label{sec:transfer2}
\begin{figure}[t]\small
\begin{center}
 \includegraphics[width=\linewidth]{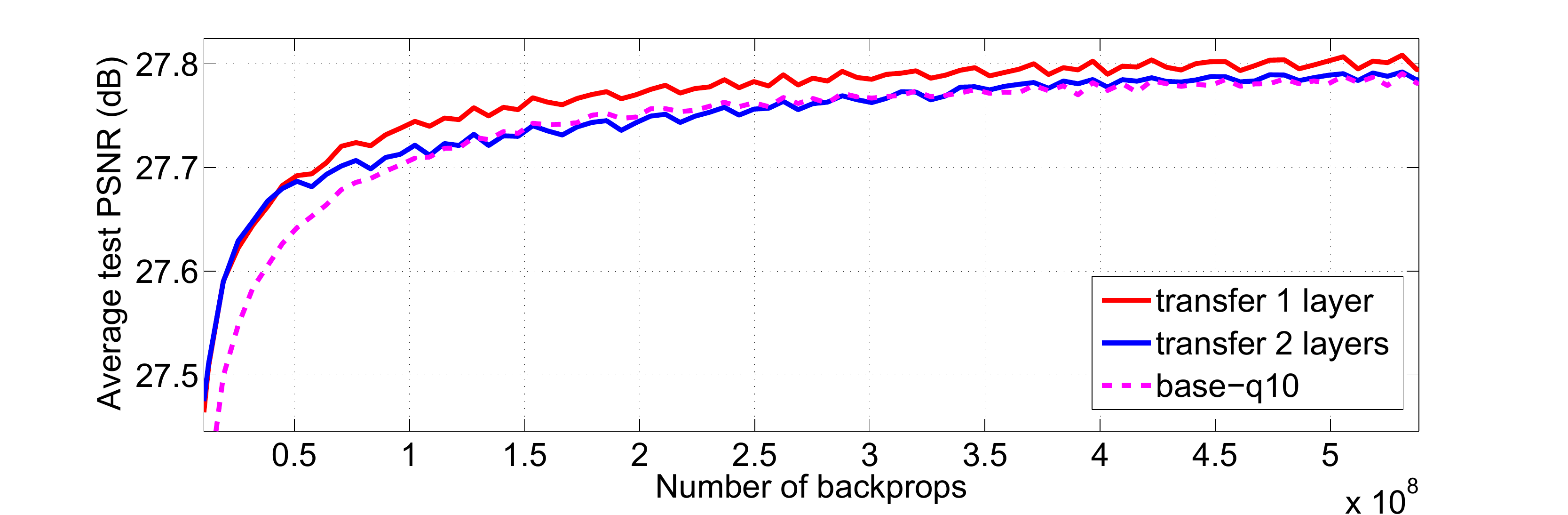}
\vskip -0.15cm
\caption{Transfer high to low quality.}
\label{fig:transfer2}
\vspace{-0.75cm}
\end{center}
\end{figure}

\begin{figure}[t]\small
\begin{center}
 \includegraphics[width=\linewidth]{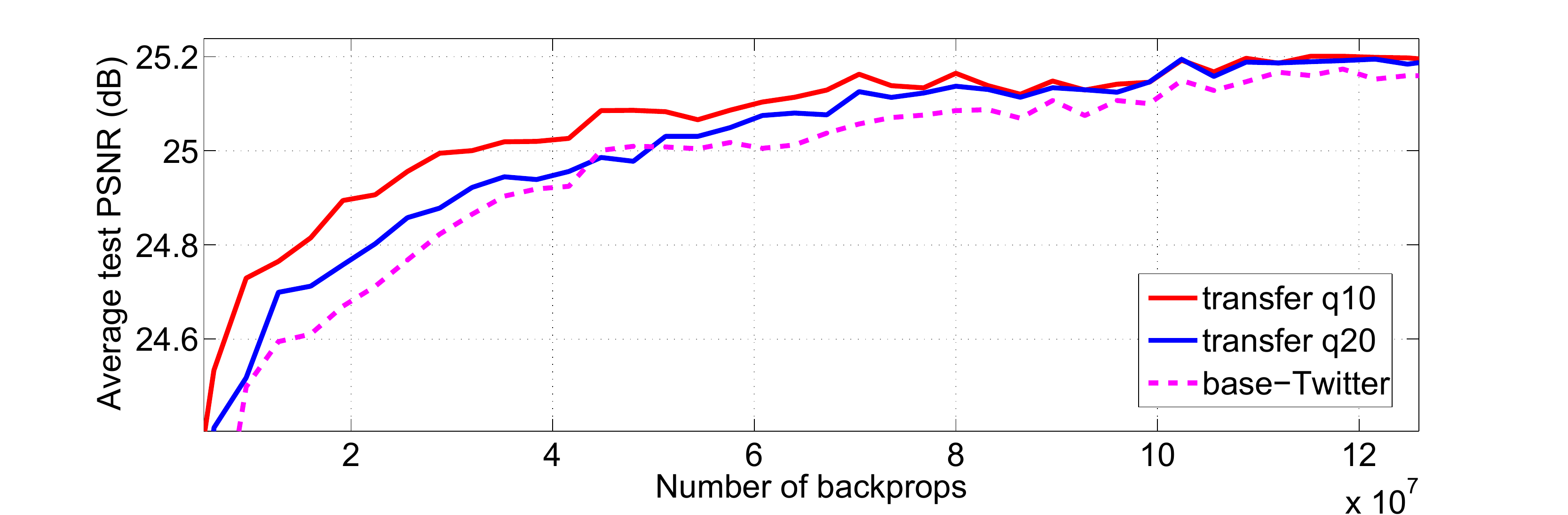}
\vskip -0.15cm
\caption{Transfer standard to real use case.}
\label{fig:transfer3}
\vspace{-0.75cm}
\end{center}
\end{figure}

Results are shown in Figure~\ref{fig:transfer2}. Obviously, the two networks with transferred features converge faster than that training from scratch. For example, to reach an average PSNR of 27.77dB, the ``transfer 1 layer'' takes only $1.54\times 10^8$ backprops, which are roughly a half of that for ``base-q10''. Moreover, the ``transfer 1 layer'' also outperforms the `base-q10'' by a slight margin throughout the training phase. One reason for this is that only initializing the first layer provides the network with more flexibility in adapting to a new dataset.
This also indicates that a good starting point could help train a better network with higher convergence speed.

\subsubsection{Transfer standard to real use case -- \textit{Twitter}}
\label{sec:transfer3}

Online Social Media like \textit{Twitter} are popular platforms for message posting. However, \textit{Twitter} will compress the uploaded images on the server-side. For instance, a typical 8 mega-pixel (MP) image ($3264 \times 2448$) will result in a compressed and re-scaled version with a fixed resolution of $600 \times 450$. Such re-scaling and compression will introduce very complex artifacts, making restoration difficult for existing deblocking algorithms (\eg~SA-DCT). However, AR-CNN can fit to the new data easily. Further, we want to show that features learned under standard compression schemes could also facilitate training on a completely different dataset.
We use 40 photos of resolution $3264 \times 2448$ taken by mobile phones (totally 335,209 training subimages) and their \textit{Twitter}-compressed version\footnote{We will share this dataset on our project page.} to train three networks with initialization settings listed in Table~\ref{tab:transfer}.

\begin{figure*}
\vskip -0.2cm
\small
\begin{center}
 \includegraphics[width=1\linewidth]{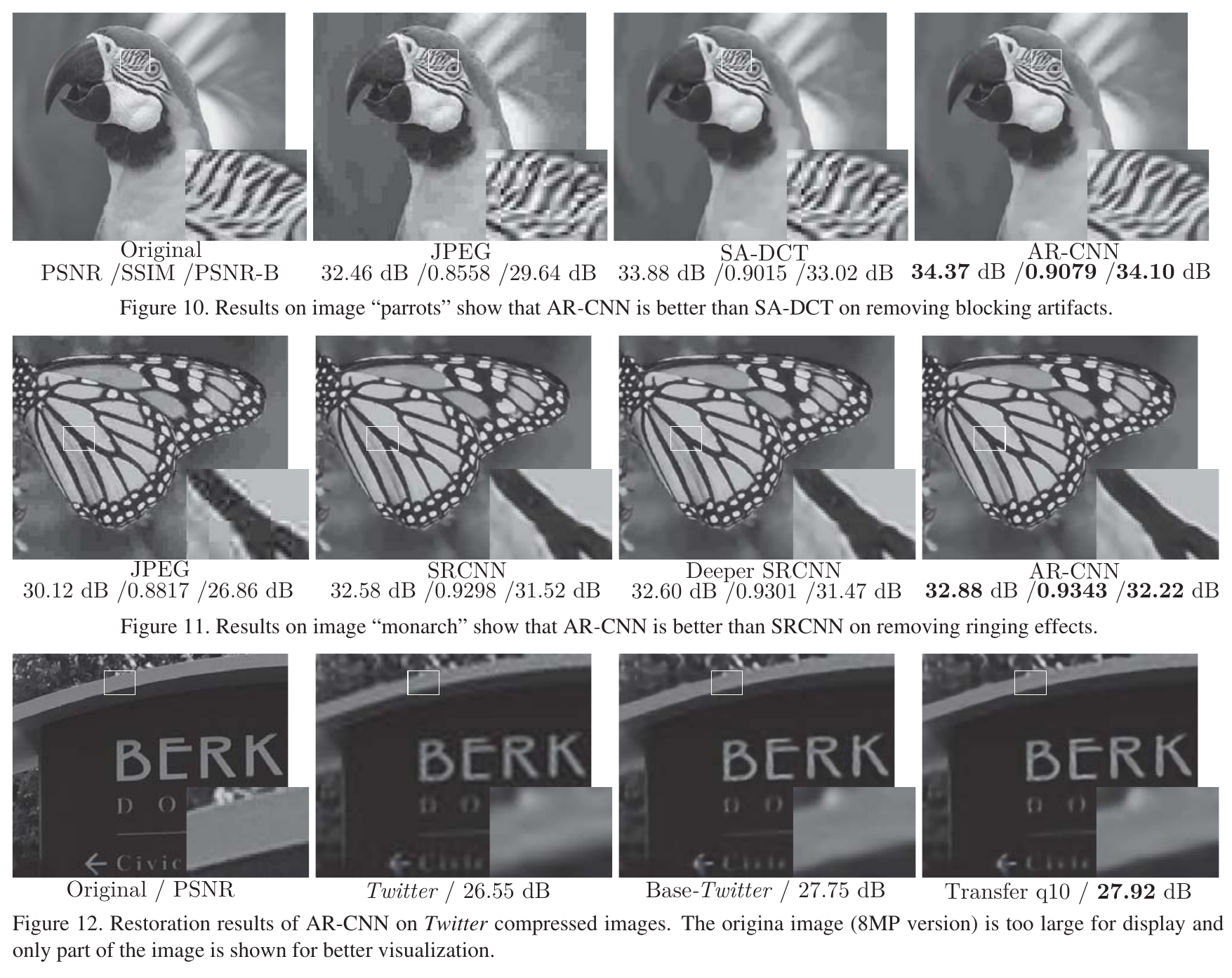}
\label{fig:last-fig}
\end{center}
\vskip -1.3cm
\end{figure*}

From Figure~\ref{fig:transfer3}, we observe that the ``transfer $q10$'' and ``transfer $q20$'' networks converge much faster than the ``base-Twitter'' trained from scratch. Specifically, the ``transfer $q10$'' takes $6\times 10^7$ backprops to achieve 25.1dB, while the ``base-Twitter'' uses $10\times 10^7$ backprops. Despite of fast convergence, transferred features also lead to higher PSNR values compared with ``base-Twitter''. This observation suggests that features learned under standard compression schemes are also transferrable to tackle real use case problems.
Some restoration results\footref{note1} are shown in Figure~\hyperlink{page.8}{\textcolor{red}{12}}. We could see that both networks achieve satisfactory quality improvements over the compressed version.

\begin{figure}[t]\small
\begin{center}
 \includegraphics[width=\linewidth]{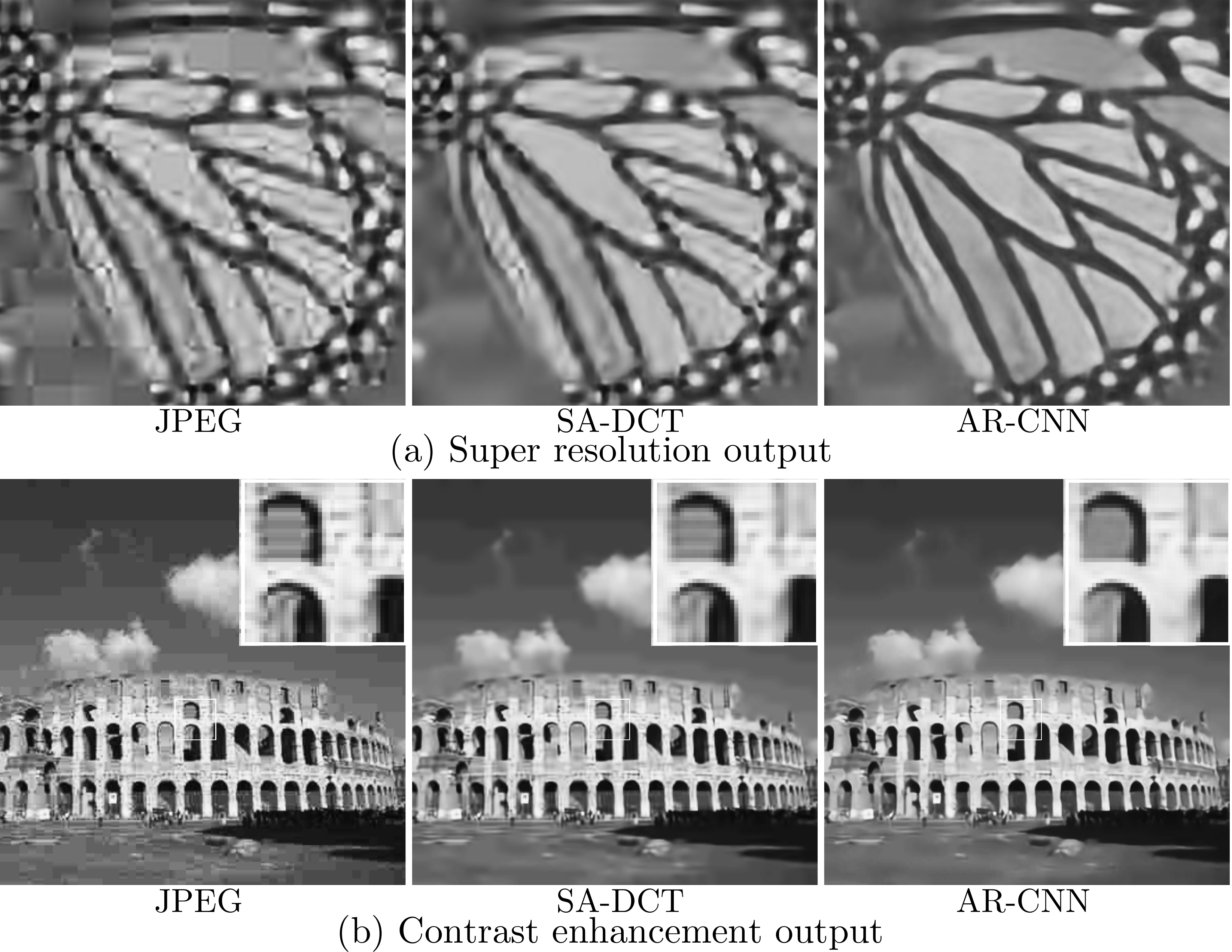}
\caption{AR-CNN can be applied as pre-processing to facilitate other low-level routines when they take JPEG images as input.}
\label{fig:application}
\vspace{-0.75cm}
\end{center}
\end{figure}

\section{Application}

In the real application, many image processing routines are affected when they take JPEG images as input. Blocking artifacts could be either super-resolved or enhanced, causing significant performance decrease. In this section, we show the potential of AR-CNN in facilitating other low-level vision studies, \ie~ super-resolution and contrast enhancement.
To illustrate this, we use SRCNN~\cite{Dong2014} for super-resolution and tone-curve adjustment~\cite{Li2014} for contrast enhancement~\cite{Dollar2013}, and show example results when the input is a JPEG image, SA-DCT deblocked image, and AR-CNN restored image. From results shown in Figure~\ref{fig:application}, we could see that JPEG compression artifacts have greatly distorted the visual quality in super-resolution and contrast enhancement. Nevertheless, with the help of AR-CNN, these effects have been largely eliminated. Moreover, AR-CNN achieves much better results than SA-DCT. The differences between them are more evident after these low-level vision processing.

\section{Conclusion}

Applying deep model on low-level vision problems requires deep understanding of the problem itself. In this paper, we carefully study the compression process and propose a four-layer convolutional network, AR-CNN, which is extremely effective in dealing with various compression artifacts. We further systematically investigate several \textit{easy-to-hard} transfer settings that could facilitate training a deeper or better network, and verify the effectiveness of transfer learning in low-level vision problems.
As discussed in SRCNN~\cite{Dong2014}, we find that larger filter sizes also help improve the performance. We will leave them to further work.

\clearpage
{\small
\bibliographystyle{ieee}
\bibliography{DCNN}
}
\end{document}